\documentclass[runningheads]{llncs}

\usepackage[mobile]{eccv}

\usepackage{eccvabbrv}

\usepackage{graphicx}
\usepackage{booktabs}
\usepackage{amsmath}
\usepackage{animate}
\usepackage{epsfig}
\usepackage{caption}
\usepackage{subcaption}
\usepackage[dvipsnames]{xcolor}
\usepackage{multirow}
\usepackage{multicol}

\usepackage[accsupp]{axessibility}  
\usepackage{wrapfig,lipsum,booktabs}

\usepackage{hyperref}

\usepackage{orcidlink}

\begin{document}

\title{Magic-Me: Identity-Specific Video Customized Diffusion} 

\titlerunning{Magic-Me}

\author{
Ze Ma\inst{1}\and
Daquan Zhou\inst{1}\and
Xue-She Wang\inst{1} \and
Chun-Hsiao Yeh\inst{2} \and
Xiuyu Li\inst{2} \and
Huanrui Yang\inst{2}\and
Zhen Dong\inst{2}\and
Kurt Keutzer\inst{2}\and
Jiashi Feng\inst{1}
} 

\authorrunning{Z.~Ma~\etal}

\institute{
ByteDance Inc.\\
\email{\{ze.ma1, daquanzhou, xueshe.wang, jshfeng\}@bytedance.com}
\and
UC Berkeley\\
\email{\{daniel\_yeh, xiuyu, huanrui, zhendong, keutzer\}@berkeley.edu}}

\maketitle

\begin{figure}[h]
\centering
\includegraphics[width=0.98\textwidth]{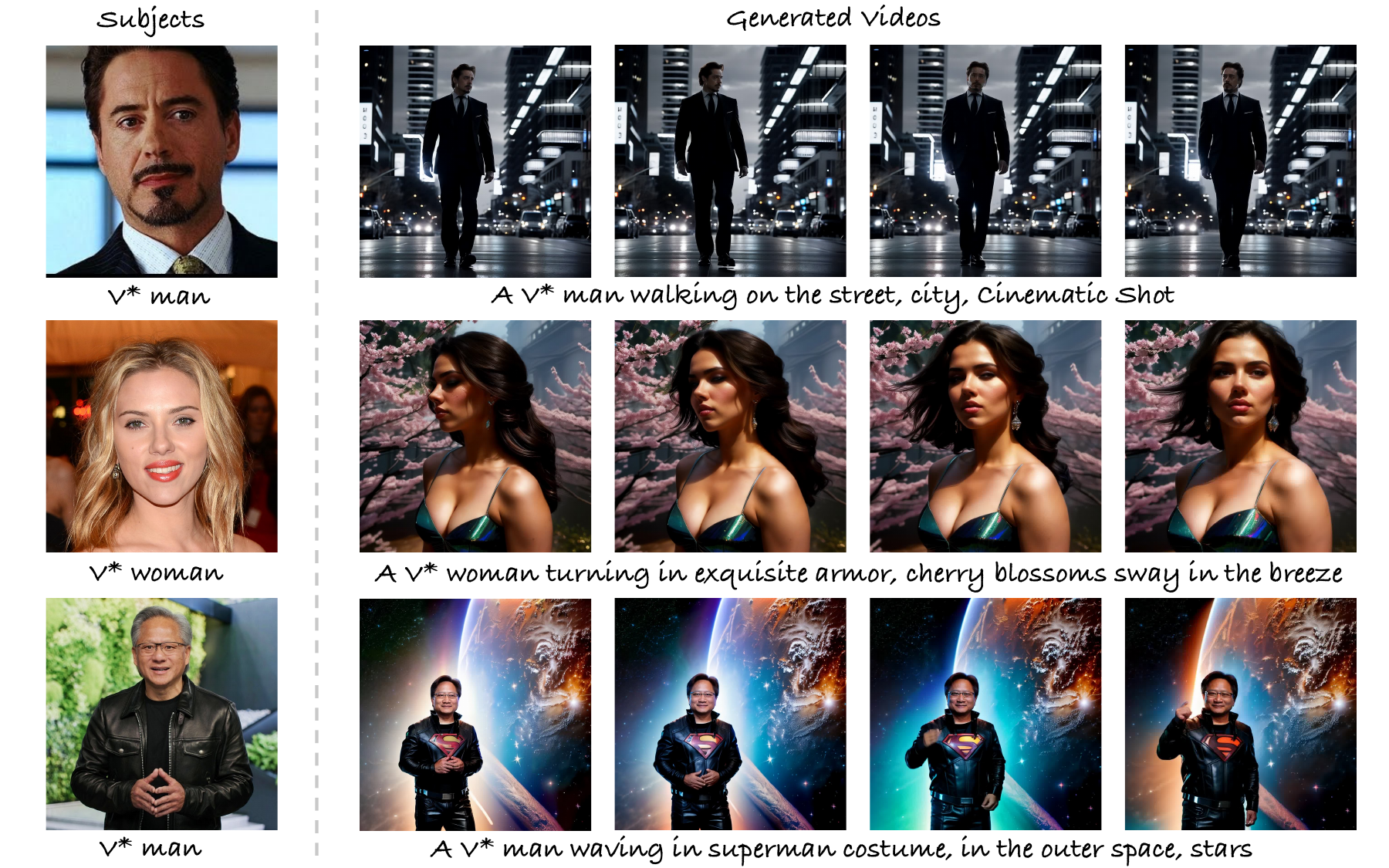}
\vspace{-3mm}
\captionof{figure}{
\textbf{ Video Custom Diffusion (VCD) results}. With just a few images of a specific identity, the proposed framework can generate temporal consistent videos aligned with the given prompt.}
\label{Figure:teaser}
\vspace{-12mm}
\end{figure}

\begin{abstract}
Creating content with specified identities (ID) has attracted significant interest in the field of generative models. In the field of text-to-image generation (T2I), subject-driven creation has achieved great progress with the identity controlled via reference images. However, its extension to video generation is not well explored. In this work, we propose a simple yet effective subject identity controllable video generation framework, termed Video Custom Diffusion (VCD). With a specified identity defined by a few images, VCD reinforces the identity characteristics and injects frame-wise correlation at the initialization stage for stable video outputs. To achieve this, we propose three novel components that are essential for high-quality identity preservation and stable video generation: 1) a noise initialization method with 3D Gaussian Noise Prior for better inter-frame stability; 2) an ID module based on extended Textual Inversion trained with the cropped identity to disentangle the ID information from the background 3) Face VCD and Tiled VCD modules to reinforce faces and upscale the video to higher resolution while preserving the identity's features. We conducted extensive experiments to verify that VCD is able to generate stable videos with better ID over the baselines. Besides, with the transferability of the encoded identity in the ID module, VCD is also working well with personalized text-to-image models available publicly. The codes are available at \url{https://github.com/Zhen-Dong/Magic-Me}. 
\vspace{-2mm}
\vspace{-2mm}

\end{abstract}

\section{Introduction}

Recent advancements in text-to-video (T2V) generation~\cite{villegas2022phenaki, yan2021videogpt, ho2022video, ho2022imagen, zhou2022magicvideo, hong2022cogvideo} have facilitated the creation of realistic animations from text descriptions. However, achieving precise control over the generated identities remains a challenge. In real-world applications, there is often a requirement to render a specific identity guided by text-described contexts, a task known as \textit{identity-specific generation}~\cite{avrahami2023bas,kumari2023multi,gal2022image}. This is crucial in scenarios such as movie production, where a specific character needs to be directed to perform particular actions.

\begin{figure}[h]
\vspace{-5mm}
\centering
\includegraphics[width=0.95\textwidth]{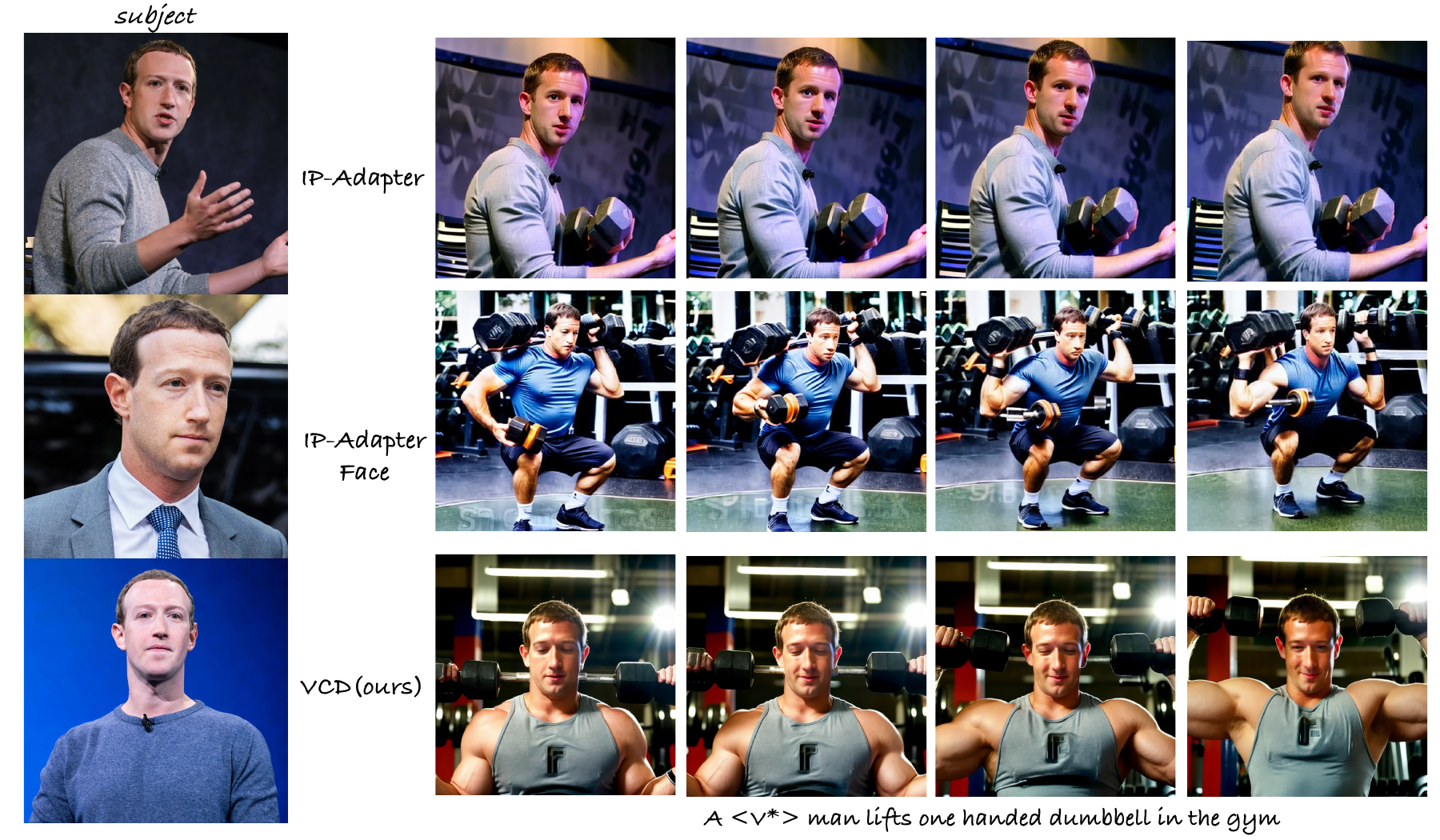}
\vspace{-3mm}
\captionof{figure}{
\textbf{Comparison between the proposed Video Custom Diffusion (VCD) method\protect\footnotemark and previous approaches}. The first two rows display results using adapters such as IP-Adapter and IP-Adapter Face\cite{ye2023ip-adapter}, and the bottom row showcases results from VCD. We observe that body-focused adapters tend to overlook the identity, thus losing alignment, whereas face-focused adapters exhibit artifacts due to conflicts in model parameter spaces when coupled with an off-the-shelf motion module~\cite{guo2023animatediff}, which is discussed in the work~\cite{gu2023mix}. Moreover, the small faces in the images appear plain and lack distinctive identity features. Our method, VCD, successfully generates stabilized videos that preserve distinctive facial characteristics of the identity.}
\label{Figure:i2v_comparison}
\vspace{-6mm}
\end{figure}

Controlling subject identity in video generation, especially for humans, remains a significant challenge. Previous works on the customization have primarily focused on styles and motions~\cite{wu2023tune} or video editing~\cite{liew2023magicedit}. While these approaches offer holistic control with conditions such as reference images~\cite{ni2023conditional}, reference videos~\cite{wu2023tune, Esser_2023_ICCV}, or depth maps~\cite{xing2023make}, they do not concentrate on identity-specific control. Moreover, as illustrated in the first two rows of Figure~\ref{Figure:i2v_comparison}, 
the direct integration of image customization modules such as IP-Adapter~\cite{ye2023ip-adapter} with the T2V method~\cite{guo2023animatediff}, struggle to generate stable videos with reasonable motions and accurate identity of the identity.

The observed failure cases underscore two key issues specific to human-focused video customization. 1) In comparison to rigid objects, humans have articulated joints that can display complex poses and movements, such as lifting, walking, and dancing, making it more challenging to maintain stable motions across frames. 2) Human movement generally involves full-body motion, but a person's distinctive characteristics are often found in the face, which occupies a relatively small area compared to the entire body. In the decoding process of VAEs, the face area on feature maps of the latent space is limited, making it difficult to render the intricate features of the face. Although existing customization methods~\cite{ye2023ip-adapter, kumari2023multi,gal2022image} can customize close-up face images, they meet difficulties in accurately rendering images with small faces and are not well-suited for videos that feature full-body motions.

In our paper, the main focus is on human-focused identity customization, where the goal is to animate a human identity with diverse motions and scenes and preserve the identity's characteristics. In the bottom row of Figure~\ref{Figure:i2v_comparison}, our method shows consistent full-body motions with identity's features, resolving the two aforementioned issues in existing approaches.

\footnotetext{Some of the samples are following the identity images used in \cite{yu2023freedom}}

To address the first issue, we analyze the exposure bias that leads to error accumulation during inference, and propose a novel 3D Gaussian Noise Prior to reconstruct the correlation across frames. This approach is training-free and enhances consistency in initialization during the inference phase. The covariance among the initialization noises for all frames is controlled by a covariance matrix. Consequently, the motions rendered are more stable in the generated videos.

For the second issue, we first build a robust ID module to balance the trade-off between preserving identity and aligning the user prompt in the generated videos. Upon analyzing existing image customization methods such as TI~\cite{gal2022image}, LoRA~\cite{hu2021lora}, and adapters~\cite{ye2023ip-adapter}, we observe that appending more parameters to the base model, as in IP-Adapter and LoRA, enhances identity preservation but makes the generated videos more susceptible to overfitting the reference images. In contrast, TI offers the greatest flexibility in generated motions and prompt alignment but exhibits poor identity recovery. To enhance TI's describing capability, we extend it from a single text token to multiple ones to encode more accurate identity information. To further improve the quality of learned identity, updates for the ID tokens are based on the area of distinct characteristics of the subject, utilizing a prompt-to-segmentation sub-module to more effectively distinguish the identity from the background. Empirical results demonstrate that the proposed ID module based on extended TI achieves a better balance between preserving identity and aligning user prompts.

Equipped with the proposed robust ID module, we integrate the ID module with a 3D Gaussian Noise Prior across three stages of Video Customized Diffusion (VCD): (1) Text-to-Video (T2V) VCD, (2) Face VCD, and (3) Tiled VCD. T2V VCD generates initial videos from the user prompt, incorporating identity characteristics in low resolution. Subsequently, during the Face VCD stage, the faces across all frames are cropped, upscaled, and regenerated with more ID-specific details. Face VCD employs a partial denoising to maintain context consistency around the face as in the initial video. Then the face videos are downscaled and reintegrated into the initial videos. Tiled VCD stage upscales and regenerates the composite video to render both the identity and the background in higher resolution. The resulting videos show more details of identity features while maintaining motion consistency across frames as shown in Figure~\ref{Figure:teaser}.

The proposed VCD framework introduces a modular approach to ID-specific video generation. The three stages of VCD reuses the same ID module to achieve generation stability in the denoising. Furthermore, based on Stable Diffusion~\cite{stablediffusion2022}, the proposed VCD can replace the base model with any domain-specific fine-tuned model stemming from the same base, and integrates other conditional inputs such as poses, depths and emotions. This offers valuable flexibility for content generation in AIGC communities such as Civitai~\cite{civitai2022} and Hugging Face~\cite{huggingface2022}, allowing non-technical users to mix and match modules independently.

Our contributions are summarized as follows:
\begin{enumerate}
    \item We introduce a novel framework, Video Custom Diffusion (VCD), dedicated to generating high-quality ID-specific videos in three cascaded stages. VCD demonstrates substantial improvement in aligning generated videos with reference images and user inputs.
    \item To render stable motions for human identity, we propose a training-free 3D Gaussian Noise Prior for video frame initialization, reconstructing inter-frame correlation and thereby improving motion consistency.
    \item To enhance the identity characteristics of human, we integrate a robust ID module with partially denoising processes in Face VCD and Tiled VCD to render more details of identities in higher resolution.
\end{enumerate}

\begin{figure}[!ht]
\vspace{-3mm}
\centering
\includegraphics[width=0.98\textwidth]{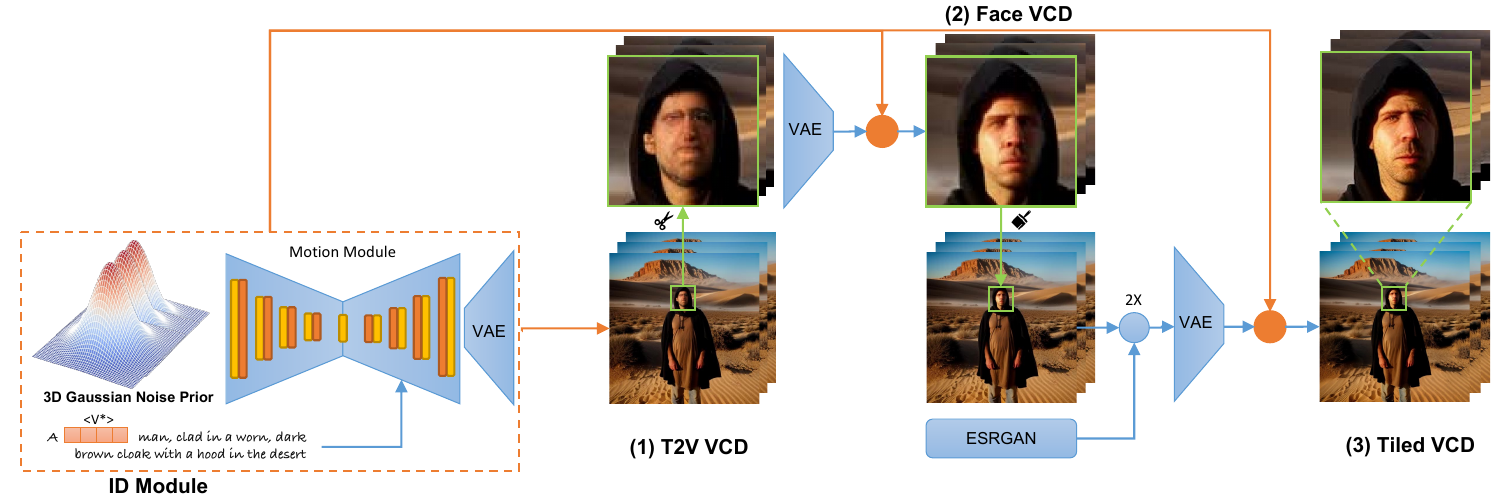}
\captionof{figure}{
{\bf Framework of ID-specific Video Generation.} The framework comprises T2V VCD, Face VCD, and Tiled VCD. The basic components ID module and 3D Gaussian Noise Prior are reused in these denoising stages.
\label{Figure:framework}
}
\vspace{-5mm}
\end{figure}

\section{Related Works}
\label{sec:formatting}
\subsection{Subject-Driven Text-to-Image Generation}
The progression of T2I diffusion models represents a remarkable stride in image generation, creating both realistic portraits and imaginative depictions of fantastical entities~\cite{rombach2022high,saharia2022photorealistic,podell2023sdxl,dalle3}. Recent efforts have spotlighted the customization of these generative models, wherein a pre-trained T2I diffusion model is employed alongside a minimal set of customized subject images, aiming to fine-tune the model and learn a unique identifier linked to the desired subject. Pioneering approaches, such as Textual Inversion~\cite{gal2022image}, adapt a token embedding to learn a mapping between token and subject images without altering the model structure, while DreamBooth~\cite{ruiz2023dreambooth} involves the comprehensive model fine-tuning to learn the concept of subject as well as preserving the capability of general concept generation. This sparked a series of subsequent works, such as NeTI~\cite{alaluf2023neural}, focusing on fidelity and identity preservation of the subject. It further extended to multi-subject generation~\cite{kumari2023multi,ma2023subject,liu2023cones,Han_2023_ICCV,bai2023integrating,gu2023mix,xiao2023fastcomposer,avrahami2023bas}, where the model is able to jointly learn multiple subjects, and compose them into a single generated image. Previous work~\cite{avrahami2023bas} introduces the masked loss for image rendering, which we adopt in the sub-module prompt-to-segmentation to further distinguish identities from the background.

\subsection{Text-to-Video Generation}

Advancing from image generation, T2V appears to be the next breakthrough in the novel applications of generative models. Compared to image generation, video generation is more challenging as it requires high computation costs to maintain long-term spatial and temporal consistency across multiple frames, needs to condition on the vague prompt of short video captioning, and lacks high-quality annotated datasets with video-text pairs. 
Early explorations utilize GAN and VAE-based methods to generate frames in an auto-regressive manner given a caption~\cite{li2018video,pan2017create}, yet these works are limited to low-resolution videos with simple, isolated motions. The next line of work adopts large-scale transformer architectures for long, HD quality video generations~\cite{villegas2022phenaki,singer2022make,hong2022cogvideo,yan2021videogpt}, yet suffering from significant training, memory, and computational costs.
The recent success of diffusion models leads a new wave of video generation with diffusion-based architectures, with pioneering work like Video Diffusion Models~\cite{ho2022video} and Imagen Video~\cite{ho2022imagen} that introduce new conditional sampling techniques for spatial and temporal video extension. MagicVideo~\cite{zhou2022magicvideo} significantly improves generation efficiency by generating video clips in a low-dimensional latent space, which is later followed by Video LDM~\cite{blattmann2023align}. Mask Diffusion augments the cross-attention module to reinforce the text control-ability \cite{zhou2023maskdiffusion}

\subsection{Video Editing}

Further advances take more control of the generated video. Tune-a-Video~\cite{wu2023tune, wu2023cvpr} allows changing video content while preserving motions by finetuning a T2I diffusion model with a single text-video pair. Text2Video-Zero~\cite{text2video-zero} and Runway Gen~\cite{esser2023structure} propose to combine a trainable motion dynamic module with a pre-trained Stable Diffusion, further enabling video synthesis guided by both text and pose/edge/images, without using any paired text-video data. More recently, AnimateDiff~\cite{guo2023animatediff} animates most of the existing personalized T2I models by distilling reasonable motion priors in the training of a motion module. 

\subsection{Image Animation}
Previous works on image animation mainly focus on extending a static image to a sequence of frames without any change of scene or modifying attributes of the character. Previous works take the subject from an image~\cite{siarohin2019animating,Siarohin_2019_NeurIPS,siarohin2021motion,wang2022latent,tao2022structure,xu2021move,zhao2022thin,dorkenwald2021stochastic} or a video~\cite{wiles2018x2face,wang2018vid2vid,wang2019fewshotvid2vid,ha2020marionette,Ni_2023_WACV}, and transfer the motion happened in another video to the subject. Our framework is able not only to animate a given frame but also to modify the attributes of the subject and change the background, all rendered in reasonable motions.

\section{Preliminaries}
\label{sec:prelimiaries}
\paragraph{Latent Diffusion Model.} Our work is based on Stable Diffusion~\cite{stablediffusion2022}, a variant of Latent Diffusion Model (LDM)~\cite{rombach2022high}. In the training, the diffusion model takes an image $x_0$ and a condition $c$ as the input and encodes $x_0$ into a latent code $z_0$ with an image encoder~\cite{kingma2013auto,isola2017image,zhang2018unreasonable}.
The latent code $z_0$ is iteratively mixed with Gaussian noise $\epsilon$ through {\it forward process}, which can be transformed into a closed form:
\begin{equation}
\label{eq:training-sample}
z_t = \sqrt{\Bar{\alpha}_t} z_0 + \sqrt{1 - \Bar{\alpha}_t} \epsilon, \epsilon \sim \mathcal{N}(0, I),
\end{equation}

where $ \Bar{\alpha}_t = \prod_{i=1}^{t} \alpha_i, \alpha_i \in (0, 1) $.

The diffusion model is trained to approximate the original data distribution with a denoising objective:
\begin{equation}
\label{eq:training-objective}
\mathbb{E}_{z_0,c,\epsilon,t} [ \left\| \epsilon_{\theta}(z_t, c, t) - \epsilon \right\|],
\end{equation}
where $\epsilon_{\theta}$ is the model prediction. In the inference, given the random Gaussian noise initialization $z_T$ and condition $c$, diffusion model performs {\it reverse process} for $t = T, \ldots, 1$ to get the encoding of sampled image $\hat{z}_0$ by the equation:
\begin{equation}
\label{eq:inference-objective}
\hat{z}_{t-1} = \frac{1}{\sqrt{\alpha_t}} \left( \hat{z}_t - \frac{1-\alpha_t}{\sqrt{1-\Bar{\alpha}_t}} \epsilon_\theta(\hat{z}_t, c, t) \right) + \sigma_t \mathbf{\epsilon},
\end{equation}
where $ \sigma_t = \frac{1 - \Bar{\alpha}_{t-1}}{1 - \Bar{\alpha}_t} \beta_t, \beta_t = 1 - \alpha_t $.

\paragraph{AnimateDiff.} We use off-the-shelf motion module AnimateDiff~\cite{guo2023animatediff} in the framework. This chosen motion module expands the network to encompass the temporal dimension. It transforms 2D convolution and attention layers into temporal pseudo-3D layers~\cite{ho2022video}, adhering to the training objective outlined in Equation~\ref{eq:training-objective}.

\section{Method}

\begin{figure*}[!t]
\vspace{-5mm}
\centering
\includegraphics[width=0.98\textwidth]{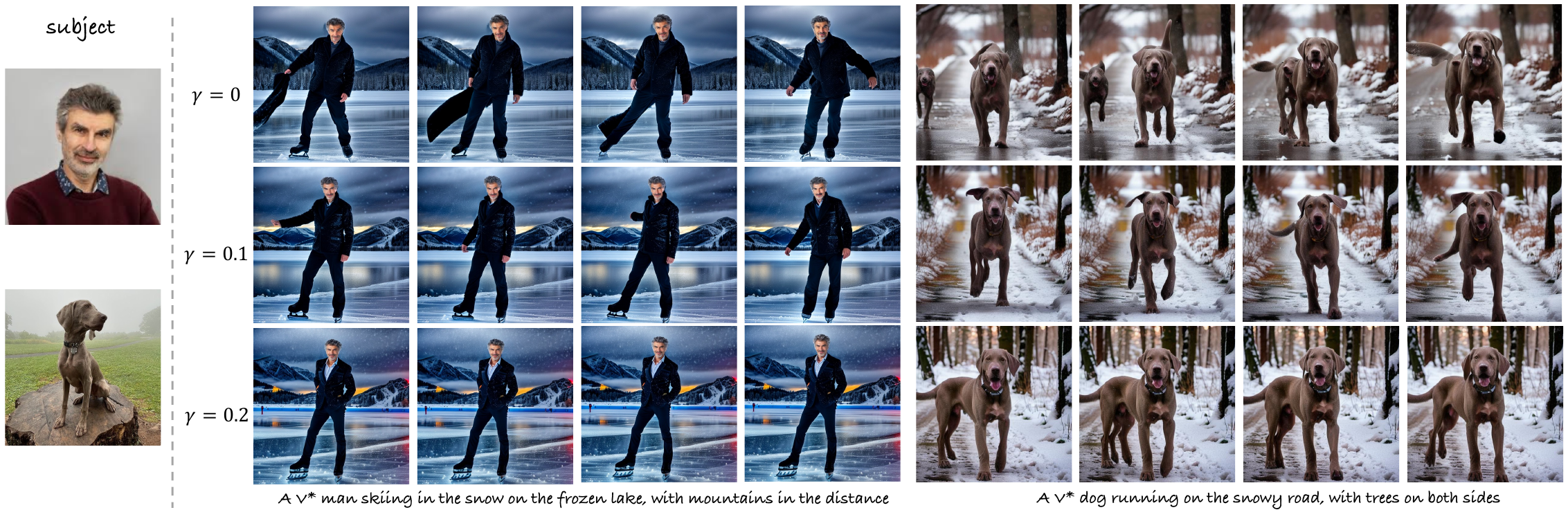}
\captionof{figure}{
{\bf Influence of 3D Gaussian Noise Prior Covariance $\gamma$.} We present results for highly-articulated subjects such as humans and animals. As the covariance hyper-parameter $\gamma$ increases, the movements across frames become more stable, and artifacts are suppressed. Setting $\gamma=0.2$ renders the video almost still.
\label{fig:gamma_ablation}
}
\vspace{-5mm}
\end{figure*}

We propose VCD framework consisting of three stages to enhance the ID features along with a 3D Gaussian Noise Prior to stabilize the human motions across the frames as illustrated in Figure~\ref{Figure:framework}. We first analyze the exposure bias in the diffusion model, and proopose 3D Gaussian Noise Prior in Section~\ref{subsec:3d_noise}. Then we discuss the trade-off between the text-alignment and ID similarity for the existing methods, and propose our ID module that make a better balance in Section~\ref{subsec:ID-module}. In the following Section~\ref{subsec:vcd}, we introduce T2V VCD that utilizes the off-the-shelf motion module of AnimateDiff~\cite{guo2023animatediff}, Face VCD to enhance the facial characteristics of the identity, and Tiled VCD that up-scale the videos to higher resolution by tiled denoising. 
\subsection{3D Gaussian Noise Prior}
\label{subsec:3d_noise}

By comparing Equation~\ref{eq:training-objective} with Equation~\ref{eq:inference-objective}, we observe that the inputs to the model $\epsilon_\theta$ differ between training and inference stages. Specifically, during training, the model $\epsilon_{\theta}(z_t, c, t)$ uses $z_t$ that is sampled from an mixture of Gaussian noise and a video, thus $z_t$ typically exhibits temporal correlation across the frames. While during inference, the model $\epsilon_\theta(\hat{z}_t, c, t)$ receives $\hat{z}_t$ that is initialized independently on each frame. The model needs to construct the temporal consistency from the stretch, a process that are not included in the training. Previously, such discrepancy are discussed in the context of exposure bias in both of Natural Language Processing and Computer Vision~\cite{ning2023input,schmidt-2019-generalization, ranzato2015sequence}, and usually leads to accumulated errors in inference. 

To mitigate this issue, we propose a training-free 3D Gaussian Noise Prior to reconstruct the temporal correlation in the initialization of $\hat{z}_T$ in the inference. For videos comprising $f$ frames, the 3D Gaussian Noise Prior samples from a Multivariate Gaussian distribution $\mathcal{N}(\mathbf{0}, \mathbf{\Sigma}_f(\gamma))$. Here, $\mathbf{\Sigma}_f(\gamma)$ denotes the covariance matrix parameterized by $\gamma \in (0, 1)$.

\begin{equation} 
\label{eq:cov_matrix}
\mathbf{\Sigma}_f(\gamma) =
\begin{pmatrix}
1 & \gamma & \gamma^2 & \cdots & \gamma^{f-1} \\
\gamma & 1 & \gamma & \cdots & \gamma^{f-2} \\
\gamma^2 & \gamma & 1 & \cdots & \gamma^{f-3} \\
\vdots & \vdots & \vdots & \ddots & \vdots \\
\gamma^{f-1} & \gamma^{f-2} & \gamma^{f-3} & \cdots & 1 \\
\end{pmatrix}.
\end{equation}

The covariance described above ensures that the initialized 3D noise exhibits a covariance of $\gamma^{|m-n|}$ at the same position between the $m$-th and $n$-th frames. The hyper-parameter $\gamma$ represents a trade-off between the stability and magnitude of the motions, as demonstrated in Figure~\ref{fig:gamma_ablation}. A lower $\gamma$ value leads to videos with dramatic movements but increased instability, while a higher $\gamma$ results in more stable motion with reduced motion amplitude.

\subsection{ID Module}
\label{subsec:ID-module}

\begin{figure}[!t]
\vspace{-0.5cm}
\centering
\begin{subfigure}[b]{0.6\textwidth}
        \centering
\includegraphics[width=\textwidth]{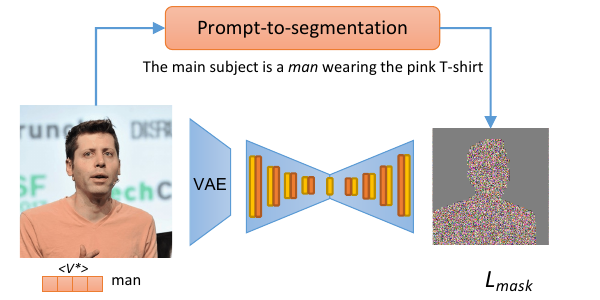}
\caption{Illustration of extended TI token training.}
\end{subfigure}
\begin{subfigure}[b]{0.35\textwidth}
        \centering
\includegraphics[width=\textwidth]{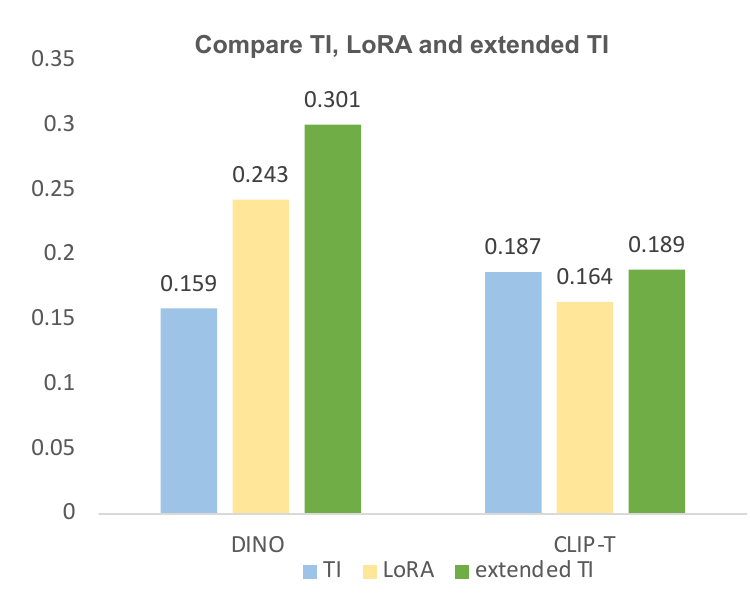}

        \caption{Compare LoRA, TI and extended TI. }

\end{subfigure}
\addtocounter{figure}{-1}
\captionof{figure}{
{ \textbf{Extended TI training.} Figure (a) shows that the extended TI are optimized against a masked subject area by prompt-to-segmentation. From the Figure (b) we can see extended TI achieves a better balance between the text alignment and identity similarity.}
\vspace{-7mm}
\label{Figure:ID-token-learning}
}
\end{figure}

\begin{figure}[!h]
\centering
\includegraphics[width=0.98\textwidth]{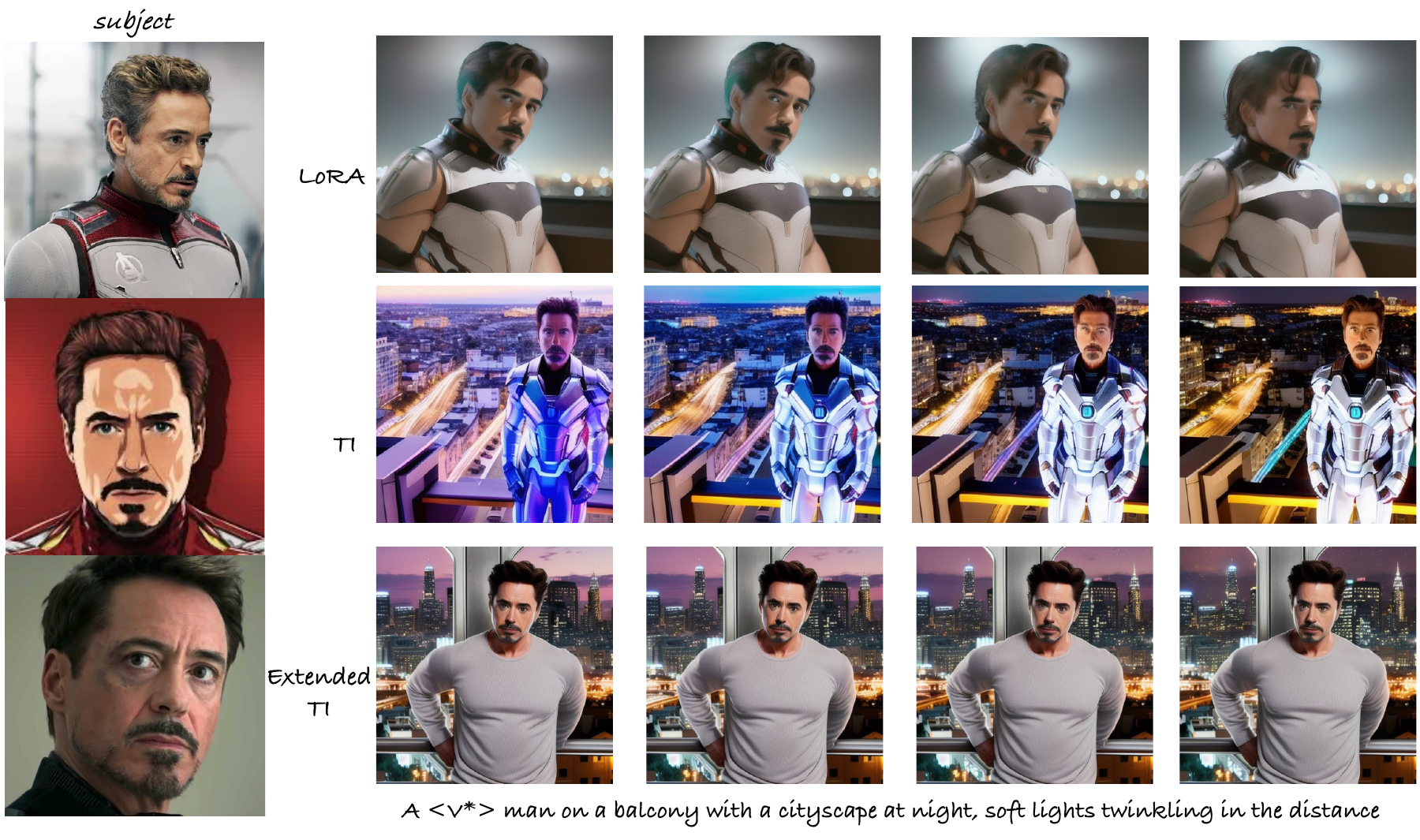}
\captionof{figure}{
\textbf{Comparison of the ID Module\cite{guo2023animatediff}\protect\footnotemark}. The first row shows the results using LoRA\cite{hu2021lora}, The middle row displays outcomes TI~\cite{gal2022image}, while the bottom row presents results from extended TI. We observe the LoRA prediction preserves more identity characteristics than TI, but is more overfitted to training images and show inferior user input alignment. With prompt-to-segmentation and sufficient ID tokens, the extended TI shows a better balance between user input alignment and similarity of the identity.}
\label{Figure:lora_ti_compare}
\vspace{-6mm}
\end{figure}

Although previous works have explored token embedding~\cite{gal2022image,voynov2023P+} and network fine-tuning~\cite{chen2023disenbooth,kumari2023multi,ruiz2023dreambooth,gu2023mix,ye2023ip-adapter} for image identity customization, few have delved into identity customization in T2V generation. 
We find that weight tuning methods such as adapters usually result in frozen characters in the video generation or conflicts with other adapters~\cite{guo2023animatediff}, while LoRA is more prone to overfit the styles of the training images thus lose the alignment of user input, as shown in Figure~\ref{Figure:lora_ti_compare}. Interestingly, we find that TI shows the best capability to align with the user prompt, though the depicted character is inferior regarding similarity.  We find that using a single token is not enough to encode the necessary information, and training ID tokens based on the entire image would introduce background interference which impede the encoding process. Thus, we make the following improvements on TI to build our ID module, as shown in Figure~\ref{Figure:ID-token-learning}.

\paragraph{Extended TI with sufficient ID tokens.} We find that with more learnable tokens,  TI is able to preserve better visual features of identities. This approach, compared to LoRA or adapters, results in better text alignment and identity resemblance. Moreover, the proposed ID module requires only 16KB of storage, a notably compact parameter space compared to the 3.6G required for parameters in Stable Diffusion or 1.7MB for SVDiff~\cite{Han_2023_ICCV} which is a compressed version of LoRA.

\paragraph{Prompt-to-segmentation.} The interference of background noise in the training is an important issue for concept customization as noted in works~\cite{chen2023disenbooth, Han_2023_ICCV}. To remove encoded background noise, we propose a straightforward yet robust method: prompt-to-segmentation. We input the prompted class information into Grounding DINO~\cite{liu2023grounding} to obtain bounding boxes. These bounding boxes are then fed into SAM~\cite{kirillov2023segany} to generate the segmentation mask of the subject. During training, we compute the loss only within the mask area.

\subsection{Video Customized Diffusion (VCD) }
\label{subsec:vcd}

The proposed VCD consists of three stages as shown in~\ref{Figure:framework}: (1) T2V VCD, a vanilla T2V process that renders videos in low-resolution, (2) Face VCD, which further enhances the characteristics on the faces, (3) Tiled VCD, that up-scales the videos to higher resolution without any degradation of the identity. All three stages reuse the same ID module and user prompt to ensure the text-alignment as well as ID similarity.
\paragraph{(1) T2V VCD} In this stage, we first utilize ID module and an off-the-shelf motion module~\cite{guo2023animatediff} to render a low-resolution videos with the user specified identity. In this process, the frames are all initialized with 3D Gaussian Noise Prior, and the results show that the generated video is well aligned with user input, but the face is blurred, as the area of feature map is small and can't recover enough details to reflect the intricate characteristics on the face.

\paragraph{(2) Face VCD.}  Face VCD regenerates the face with more identity characteristics without changing the video's resolutions. It first utilizes the prompt-to-segmentation sub-module to detect the bounding box of faces and segment the pixels on the face. Bounding boxes of faces are diluted to include the background around the faces and give more contexts for the ID module. The bounding boxes are then cropped and stacked to make a new face video. Face VCD up-scales the face video, and mixes it with partial Gaussian noise as in ~\cite{wu2023tune}. The ID module denoises the mixture of face video to  recover more details of the faces. After partially denoising, the regenerated faces have more distinctive characteristics and they are down-scaled to the original resolution so that they can be pasted the initial videos. From the Figure~\ref{Figure:framework} we can see even after the downsampling, the recovered faces in this stage still shows more characteristics and natural existence in the contextual background. 

\paragraph{(3) Tiled VCD.} The output after Face VCD is still limited in resolution (512x512). Tiled VCD can further upscale the video while preserving the identity. The videos are first up-scaled to 1024x1024 by ESRGAN~\cite{ledig2017photo,wang2021real}. However, during the upscaling, ESRGAN has no prior knowledge about the user prompt or the specified identity. Thus, the up-scaled need to be denoised using ID module again. We segmented the up-scaled video into 4 tiles, each of them occupying 512x512 pixels as the original video. Then we add partial 3D Gaussian Noise Prior to the video for ID module to denoise. Each tile is partially denoised with the same ID module, with the details of identity rendered in higher resolution, and stitched together. Both of the identity and background are more clear in this stage.

\begin{figure}[!t]
    \vspace{-3mm}
    \centering
    \includegraphics[width=1.0\textwidth]{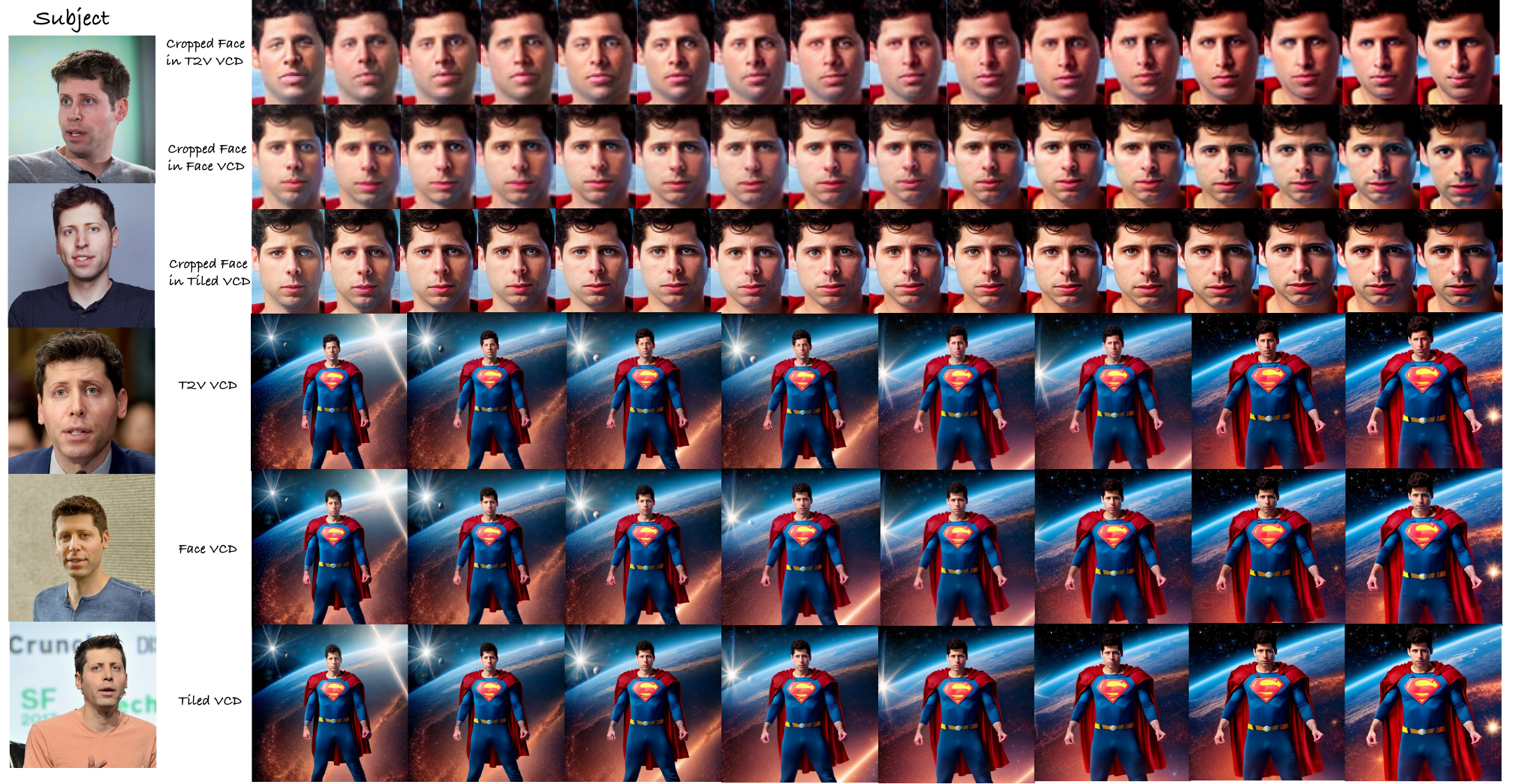}
\caption{\textbf{VCD Qualitative Results.} We display the zoomed-in faces after each stage. From the figure, it is evident that the faces are significantly enhanced with the identity's characteristics following Face VCD and Tiled VCD interventions. The prompt used is {\it a V* man floating in a Superman costume, in outer space, among stars}.}
    \label{fig:vcd_qualitative}
    \vspace{-3mm}
\end{figure}

\section{Experiments}
In this section, we presents the experimental evaluation of our method. We begin with an overview of the implementation details. Following this, we present qualitative results demonstrating the animation of various identities, controllable generation, as well as quantitative results. Finally, we conclude with ablation studies.

\subsection{Implementation Details}
\paragraph{Training.} 

Unless specified otherwise, the ID module is trained using Stable Diffusion 1.5 and employed with Realistic Vision during inference. Applying it directly to Stable Diffusion 1.5 for video generation coupled with AnimateDiff~\cite{guo2023animatediff} results in inferior videos. We use 4 text tokens for extended TI. We set the learning rates for extended token tokens at 1e-3. The batch size is fixed at 4. Each identity's ID module undergoes 200 optimization steps during training. For the motion module, we adjust the $\gamma$ in Equation~\ref{eq:cov_matrix} to 0.1.  We denoise 80\% in the Face VCD while 20\% in Tiled VCD.

\paragraph{Datasets.} To validate the effectiveness of our proposed VCD framework, we collected training images of 24 characters from CustomConcept101~\cite{kumari2023multi} as well as from the internet, ensuring a diverse representation of humans and animals. For each subject, we tasked GPT-4V with creating 25 prompts for animations against various backgrounds. For evaluation purposes, the model generates four videos for each prompt, using different random seeds. This process results in a total of 2400 videos.

\paragraph{Evaluation Metrics.} We evaluate the generated videos from three perspectives. (i) Identity alignment: The visual appearance of the generated identity should align with that in the reference images. We utilize CLIP-I~\cite{radford2021learning} and DINO~\cite{caron2021emerging} to compute the similarity score between each pair of video frames and reference images. (ii) Text-alignment: The text-image similarity score is calculated in the CLIP feature space~\cite{hessel2021clipscore}. (iii) Temporal-smoothness: We assess temporal consistency in the generated videos by computing  CLIP scores for all pairs of consecutive video frames. It's important to note that temporal smoothness is influenced not only by the content consistency between consecutive frames but also by the motion's magnitude. Therefore, it is important to consider text alignment, identity similarity, and temporal smoothness together when comparing results.

\paragraph{Baselines.} Due to the lack of identity-specific T2V methods, we compare our method with direct composition of concept customization methods with AnimateDiff, such as CustomDiffusion~\cite{kumari2023multi}, Textual Inversion (TI)~\cite{gal2022image}, IP-Adapter~\cite{ye2023ip-adapter}, and LoRA~\cite{ryu2023low, hu2021lora}, all combined with a 3D Gaussian Noise Prior. While recent advancements have introduced more novel customization methods for multi-identity customization, such as those found in~\cite{gu2023mix, Han_2023_ICCV,chen2023videodreamer,wang2024customvideo }, the integration with such methods could be left for future work.

\subsection{Qualitative Results}
\vspace{-3mm}
\begin{figure}[!h]
    \vspace{-3mm}
    \centering
    \includegraphics[width=1.0\textwidth]{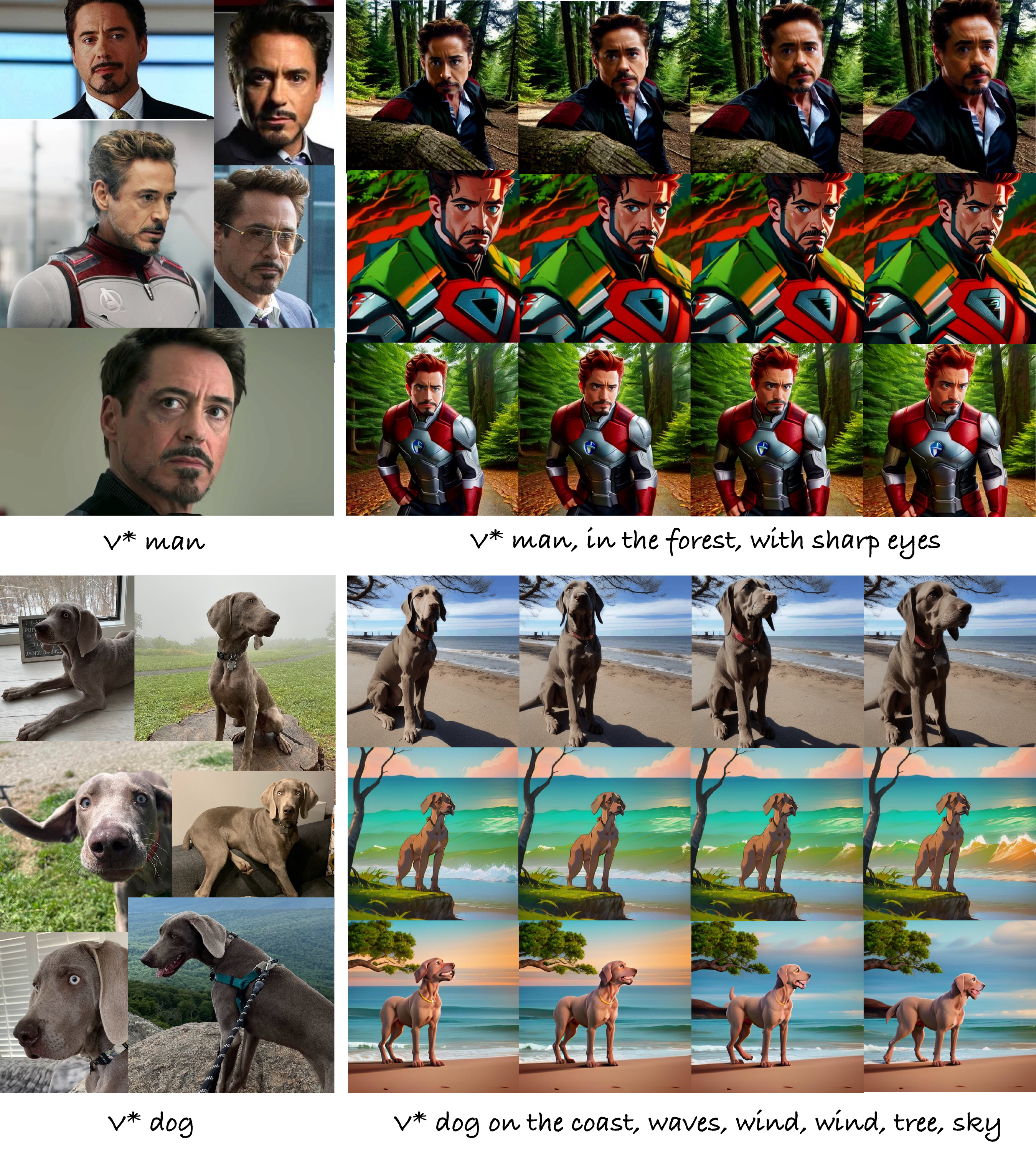}
    \caption{\textbf{Compatibility with Personalized Base Models.} We list results from Realist Vision~\cite{realisticvision2023}, ToonYou~\cite{toonyou2023} and RCNZ Cartoon 3D~\cite{rcnzcartoon2023} for each subject.}
    \label{fig:personalized_qualitative}
    \vspace{-5mm}

\end{figure}

In this section, we first present the generated results, highlighting zoomed-in cropped faces in Figure~\ref{fig:vcd_qualitative} at each stage to showcase the effectiveness of our proposed framework. We observe that after the initial text-to-video (T2V) stage, faces appear plain and lack distinguishing features. However, following the Face VCD stage, even without any resolution changes on the displayed faces, they begin to closely resemble the specified identity. After Tiled VCD, which upscales the video clip as a whole, faces are rendered with enhanced details. Furthermore, we demonstrate that our proposed VCD framework and ID module can effectively render identities on different personalized models derived from the same SD1.5, as shown in Figure~\ref{fig:personalized_qualitative}.

\begin{table}[!t]

\centering
\caption{\textbf{Quantitative comparison with baseline models.} The best score is highlighted in bold. The symbol $^{\uparrow}$ indicates that a higher score implies greater relevance. The proposed ID module achieves an optimal balance between video consistency and image alignment. \it SD: Stable Diffusion, RV: Realistic Vision, TI: Textual Inversion, DB: DreamBooth, IA Face: IP-Adapter Face.}

\label{table:quantitative}
\resizebox{\columnwidth}{!}{%

\begin{tabular}{lccccccc}
\hline
 & SD~\cite{stablediffusion2022}& RV~\cite{realisticvision2023} & TI~\cite{gal2022image} & LoRA~\cite{hu2021lora}    & DB~\cite{ruiz2023dreambooth} &IA Face~\cite{ye2023ip-adapter} & Ours\\ \hline
DINO $^{\uparrow}$                  &   0.267     &    0.345     &     0.352       &  0.522 &   0.530  & 0.455 &   {\bf 0.585}    \\
CLIP-I $^{\uparrow}$                  & 0.595     &    0.624      &    0.701        & 0.722  &   0.699  & 0.610  &  {\bf 0.747}   \\   
CLIP-T$^{\uparrow}$                 & 0.265  &   0.275       &     0.271       & 0.238 &  0.222    & 0.250 &\bf{0.315}    \\
CLIP-I Temp Consist$^{\uparrow}$   & 0.729  &    0.758      &      0.796      & 0.810 &    0.734   & 0.771 &{\bf 0.835}
\\ \hline
\end{tabular}
}
\vspace{-2mm}
\end{table}

\begin{table}[!t]
\centering
\caption{\textbf{Influence of 3D Gaussian Noise Prior}. 
We experiment with different level covariance in 3D Gaussian Noise Prior, with higher $gamma$, video temporal consistency improves while text alignment is decreasing. When $gamma$ is higher than 0.2, the differences between frames are minimal, as illustrated in Figure~\ref{fig:gamma_ablation}.
}

\begin{tabular}{ccccc}
\hline
            & DINO$^{\uparrow}$ & CLIP-I$^{\uparrow}$ & CLIP-T$^{\uparrow}$  &  \begin{tabular}[c]{@{}c@{}}CLIP-I Temp\\ Consist$^{\uparrow}$\end{tabular} \\ \hline
$\gamma$ = 0       &       0.560                &     0.745                & 0.310 &  0.797             \\
$\gamma$ = 0.1  &     \textbf{0.585}         &      0.747          &   \textbf{0.315 }                 &          0.835       \\
$\gamma$ = 0.2   &    0.583          &      \textbf{0.816}          &       0.280   &   \textbf{0.861}               \\
\hline
\end{tabular}

\label{table:gamma_ablation}
\vspace{-2mm}
\end{table}

\subsection{Quantitative Results}
We present the quantitative results in Table~\ref{table:quantitative}. Initially, we evaluate two pre-trained models: Stable Diffusion (SD) and Realistic Vision. Realistic Vision, a community-developed model fine-tuned on SD, shows promising results in generating realistic images. As indicated in Table~\ref{table:quantitative}, Realistic Vision generally outperforms SD, leading us to adopt it as the base model when possible. However, for models like DreamBooth, which involve fine-tuning all weights in the UNet, replacing the base model weights is not feasible. Its performance is generally inferior to the others, highlighting the limitations of extensive fine-tuning. We also compare with LoRA, TI and IP-Adapter Face in the table. As vanilla IP-Adapter model can only animate the image and can't align with user input, we compare with IP-Adapter Face here.

\subsection{Controllable Generation}
\vspace{-2mm}
\begin{figure}[!t]
\vspace{-2mm}
\centering
\includegraphics[width=0.98\textwidth]{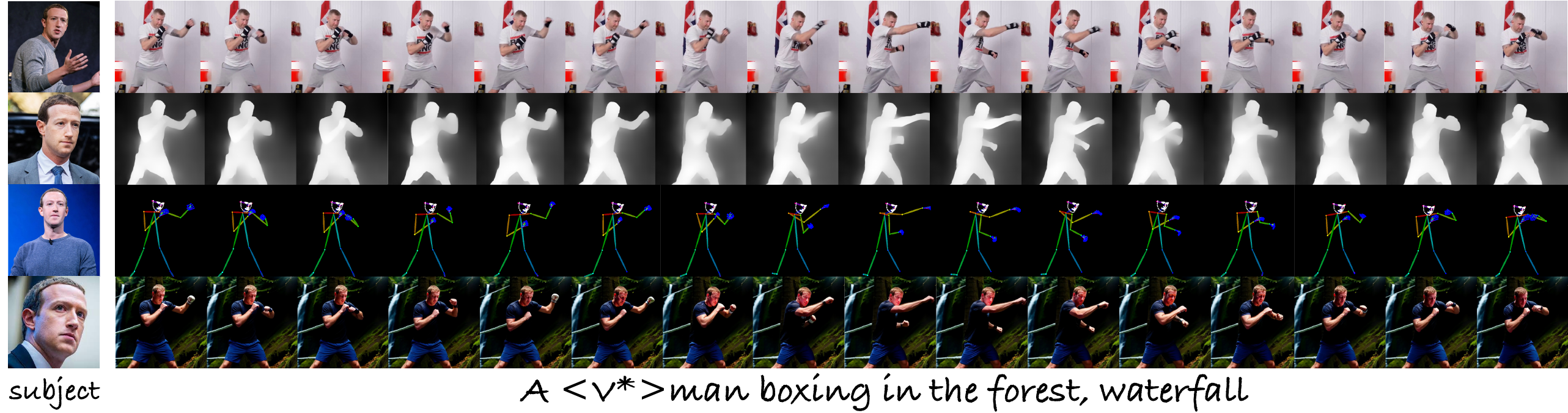}
\vspace{-2mm}
\captionof{figure}{
\textbf{VCD with ControlNet}. Our method can be integrated with conditional inputs such as depth maps and poses using ControlNet to achieve controllable video generation.}
\label{Figure:controlnet}

\end{figure}

\begin{figure}[!t]
\centering
\includegraphics[width=0.98\textwidth]{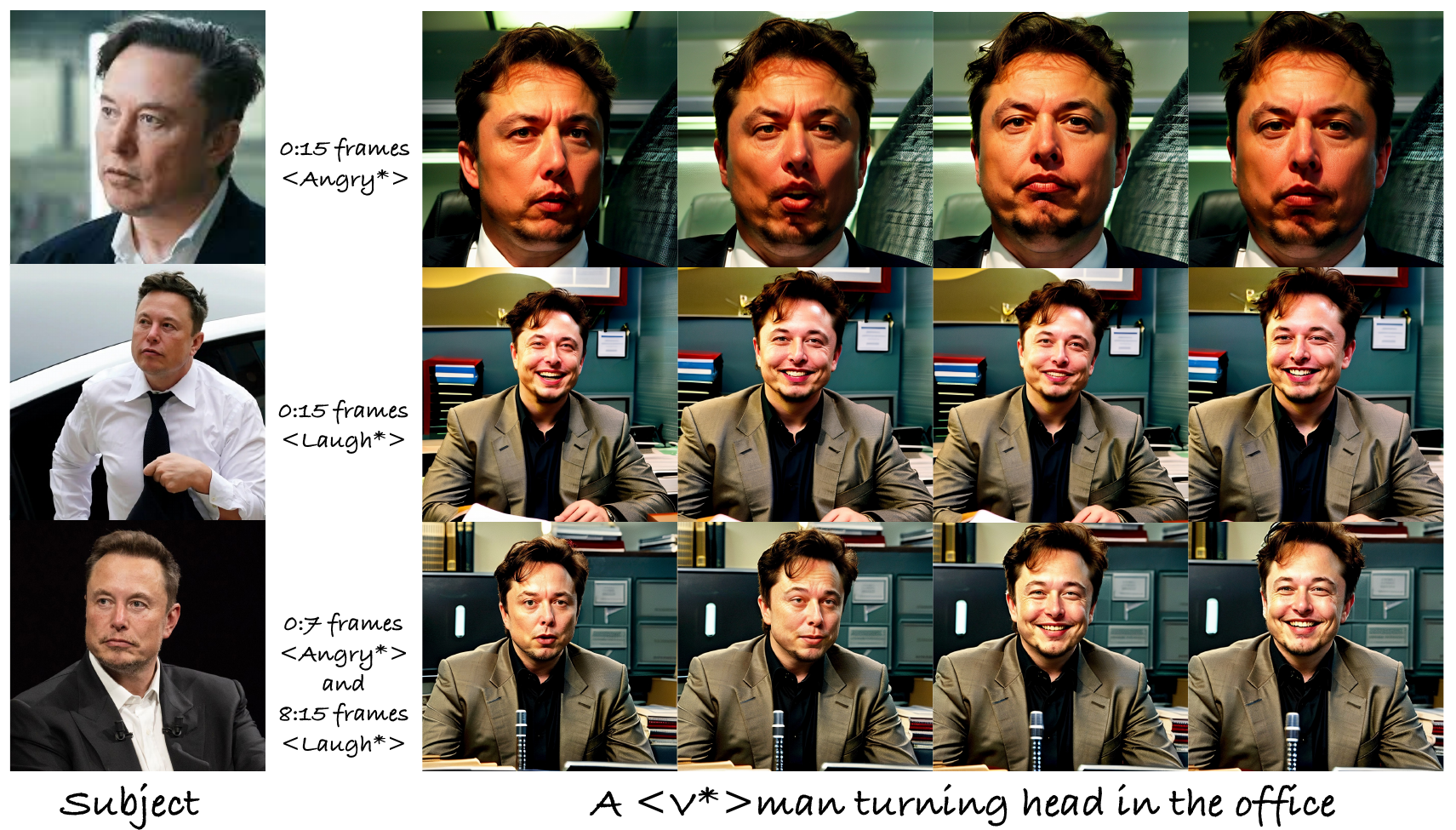}
\vspace{-2mm}
\captionof{figure}{
\textbf{VCD with Controllable Emotions}. To demonstrate the compatibility of the VCD framework, we present results featuring controllable emotions on human faces using embedding. In the first two rows, videos are generated with a consistent prompt, while the last row displays a transition of emotions, employing an angry motion for the first 8 frames followed by a laughing emotion for the remaining 8 frames.}
\label{Figure:emotion}

\end{figure}
We demonstrate that our framework supports controllable generations, as illustrated in Figure~\ref{Figure:controlnet} and Figure~\ref{Figure:emotion}. In Figure~\ref{Figure:controlnet}, where vanilla VCD faces challenges in generating sudden movements such as boxing, we utilize ControlNet with reference motion to accurately render the identity in the desired poses. Furthermore, traditional conditional inputs fail to control characters' expressions, a critical aspect for applications in the film industry. In Figure~\ref{Figure:emotion}, we present controllable emotion generation and employ multi-prompt techniques across frames to facilitate the transition of emotions.
\vspace{-2mm}

\vspace{-2mm}
\subsection{Ablation Study}

\begin{table}[t]
\centering
\caption{\textbf{Ablation study for three stages of VCD}. From the table, we can see each stage of VCD has increased the similarity between generated identities and reference ones as indicated by DINO.
}

\begin{tabular}{cccccc}
\hline
            & DINO$^{\uparrow}$ & CLIP-I$^{\uparrow}$ & CLIP-T$^{\uparrow}$ &  \begin{tabular}[c]{@{}c@{}}CLIP-I Temp\\ Consist$^{\uparrow}$\end{tabular} \\ \hline
After T2V VCD      &       0.539       &     0.721           &     0.321                &  0.826             \\
After Face VCD  &    0.551          &      \textbf{0.752}          &       \textbf{0.322}      &  0.825               \\
After Tiled VCD  &     \textbf{0.585}         &      0.747          &    0.315                  &          \textbf{0.835}       \\

\hline
\end{tabular}
\label{table:stage_ablation}
\end{table}

To validate the impact of proposed 3D Gaussian Noise Prior, we conduct a detailed ablation study  as shown in Table~\ref{table:gamma_ablation}. In the study, 3D Gaussian Noise Prior is crucial for video smoothness. Increasing $gamma$ bring more temporal consistency while losing the text alignment. We conduct ablation study for three stages of VCD as in Table~\ref{table:stage_ablation}, and find that each stage could improve the identity similarity.

\section{Limitations and Future Works}
Our proposed framework has several areas to improve. First, it struggles when we try to make videos with several different identities, each with its own special token embedding. The resulting video degrades especially when these characters have to interact with each other. Second, the proposed frameworks are limited by the capacity of the motion module. Given the motion module that only generates short-time videos, it's not easy to extend the length of the videos while keeping the same consistency and fidelity.  We need to work on making the system capable of handling multiple identities that interact with each other,  and ensuring it can keep up the quality in longer videos. 

\section{Conclusion}
 In this paper, we introduces Video Custom Diffusion (VCD), a framework designed for subject identity controllable video generation. By focusing on the encoding of identity information and frame-wise correlation, VCD paves the way for producing videos that not only maintain the subject's identity across frames but do so with stability and clarity. Our novel contributions, including the ID module for precise identity disentanglement, the T2V VCD module for enhanced frame consistency, and the Face VCD and Tiled VCD modules for improved video quality, collectively establish a new standard for identity preservation in video content. The extensive experiments we conducted affirm VCD's superiority over existing methods in generating high-quality videos that preserve the subject's identity. Furthermore, the adaptability of our ID module to work with existing text-to-image models and controllable generations enhances VCD's practicality, making it versatile for a broad range of applications.


%
%
\bibliographystyle{splncs04}
\bibliography{main}
\end{document}